# Accuracy of the Uzbek stop words detection: a case study on "School corpus"

Khabibulla Madatov [1], Shukurla Bekchanov[1] and Jernej Vičič [2,3]

[1] *Urgench state university, 14, Kh. Alimdjan str, Urgench city, 220100, Uzbekistan*
[2] *Research Centre of the Slovenian Academy of Sciences and Arts, The Fran Ramovš Institute, Novi trg 2, 1000 Ljubljana, Slovenija*
[3] *University of Primorska, FAMNIT, Glagoljaska 8, 6000 Koper, Slovenia*

**Abstract**
Stop words are very important for information retrieval and text analysis investigation tasks of natural language processing. Current work presents a method to evaluate the quality of a list of stop words aimed at automatically creating techniques. Although the method proposed in this paper was tested on an automatically-generated list of stop words for the Uzbek language, it can be, with some modifications, applied to similar languages either from the same family or the ones that have an agglutinative nature. Since the Uzbek language belongs to the family of agglutinative languages, it can be explained that the automatic detection of stop words in the language is a more complex process than in inflected languages. Moreover, we integrated our previous work on stop words detection in the example of the "School corpus" by investigating how to automatically analyse the detection of stop words in Uzbek texts. This work is devoted to answering whether there is a good way of evaluating available stop words for Uzbek texts, or whether it is possible to determine what part of the Uzbek sentence contains the majority of the stop words by studying the numerical characteristics of the probability of unique words. The results show acceptable accuracy of the stop words lists.

## 1. Introduction

The application of Natural Language Processing (NLP) tasks in real-life scenarios are getting more frequent than ever before, and there is huge research getting involved with different approaches to enhance the quality of such tasks. An important aspect of many NLP tasks that make use of tasks, such as information retrieval, text summarization, context-embedding, etc., relies on a task of removing unimportant tokens and words from the context under focus. Such data are known as stop words. Therefore, it is desired that some automatic method should be developed to identify stop words that either make no change in the meaning of the context (or do very little) and remove them. from the context.

In this work, we are addressing the problem of automatic detection of stop words for the low-resource agglutinative Uzbek language, and evaluate the proposed methods. The existing literature that deal with stop words removal task for the Uzbek language [7] [8] [10] focus on the creation process, the importance, as well as the availability of the proposed data, leaving a gap for further investigation, which we discuss in this paper.

The scientific term "stop words" is popular in the field of natural language processing, and its definition we focus in this work is as follows: If the removal of those words from the text not only does

---

not change the context meaning but also leaves the minimum number of words possible that can still hold the meaning of the context, then such words can be called stop words for this work.

For instance, the following examples are shown to better explain what words would be considered in given sentences, and what the final context would become after removing those stop words:
- ***"Men bu maqolani qiynalib yozdim".*** (I wrote this article with difficulty). After removing the stop words ("*men*", "*bu*", "*qiynalib*") the context becomes: ***"Maqolani yozdim"***.(I wrote the article.);
- ***"Har bir inson baxtli bo'lishga haqlidir"*** (Every person has the right to be happy). After removing the stop words ("*har* ", "*bir*"), the context becomes: ***"Inson baxtli bo'lishga haqlidir"*** (Person has right to be happy).

Such definition is an extension of the traditional definition of stop words by including more words than the actual expectations but still including the traditional stop words.

The Term Frequency - Inverse Document Frequency (TF-IDF) method [15] was used to detect stop words in Uzbek texts. TF-IDF is a numerical statistic that is intended to reflect how important a word is to a document in a corpus, the method acknowledges words with the lowest TF-IDF values as less important to the semantic meaning of the document and proposes these words as stop word candidates.

In our previous work[8], we discuss the methods and algorithms for automatic detection and extraction of Uzbek stop words from previously collected text forming a new corpus called the "School corpus". The stop words detection method based on TF-IDF was applied to the aforementioned corpus collected from 25 textbooks used for teaching at primary schools of Uzbekistan, consisting of 731,156 words, of which 47,165 are unique words. To perform our technique, for each word from the set of unique words, its frequency was determined (the number of occurrences in the texts of the School corpus), and the inverse document frequency IDF(word) = ln(n/m) where n = 25 – number of documents and m is the number of documents, containing the unique word among 25 documents.

The existing fundamental papers that deal with stop words in general, let alone for the Uzbek language, barely address the quality of the automatically detected list of stop words. This statement also applies to our previous work, where a preliminary manual expert observation of a part of the lists (only unigrams) was done. To the authors' knowledge, there was no in-depth observation of the accuracy of the automatically constructed lists of stop words for agglutinative languages. For instance, [7][8][9][10] are mostly focusing on Uzbek texts' stop words and methods for automatic extraction of stop words. But none of them discusses the accuracy of the presented methods. The article is devoted to answering whether there is a good way of evaluating available stop words for Uzbek texts, or whether it is possible to determine what part of the Uzbek sentence contains the majority of the stop words by studying the numerical characteristics of the probability of unique words.

The words were sorted by the TF-IDF value in descending order and the lowest 5 percent of them were tagged as stop words. We used this method to automatically detect stop words in the corpus [8]. Using this information, the article focuses on the followings:
- To create a probability distributions model of the TF-IDF of unique words in order to determine the position of stop words along with the corpus;
- To establish the accuracy of the detection method for stop words;
- To conclude on automatic position detection of stop words for the given text.

The rest of the paper is structured as follows: We start by explaining the related works in the field of stop word removal, as well as the Uzbek language itself in Section 2, followed by the main methodology of the paper in Section 3, which includes the creation of probability distribution law of TF-IDF of unique words (Section 3.1), the numerical characteristics of the probability of unique words (Section 3.2), and the evaluation of the created method using a small selected chunk (Section3.3). The accuracy of the method for automatic detection of stop words in Uzbek texts, which is based on TF-IDF, is presented in Section 4. The last section of the paper presents conclusions and future work (Section 5).

## 2. Related works

Uzbek language belongs to the family of Turkic languages. There has been some research on the Uzbek language mostly in the last few years. Most of the research done on Turkic languages can be applied to the Uzbek language as well, using cross-lingual learning and mapping approaches, alongside some language-specific additions. The paper [1] presents a viability study of established techniques to align monolingual embedding spaces for Turkish, Uzbek, Azeri, Kazakh, and Kyrgyz, members of the Turkic family which is heavily affected by the low-resource constraint.

Several authors present experiment and propose techniques for stopwords extraction from text for agglutinative languages such as [2] that bases the stopword detection problem as a binary classification problem and the evaluation shows that classification methods improve stopword detection with respect to frequency-based methods for agglutinative languages but fails for English. Ladani and Desai [5] present an overview of stopwords removal techniques for Indian and Non-Indian Languages. Jayaweera et al. [2] proposes a dynamic approach to find Sinhala stopwords, the cutoff point is subjective to the dataset. Wijeratne and de Silva [17] collected the data from patent documents and listed the stopwords using term frequency. Rakholia et al. [14] proposed a rule-based approach to detect stopwords for the Gujarati language dynamically. They developed 11 static rules and used them to generate a stopword list at runtime. Fayaza et al. [1] presents a list of stopwords for Tamil language and reports improvement in text clustering using removal.

The paper **¡Error! No se encuentra el origen de la referencia.** provides the first annotated corpus for polarity classification for the Uzbek language. Three lists of stop words for the Uzbek language are presented in [7] that were constructed using automatic detection of stop words by applying algorithms and methods presented in [8]. Paper [9] focuses on the automatic discovery of stop words in the Uzbek language and its importance. Articles [12] and [13] are also mainly concentrated on the creation of stop words in Uzbek.

Matlatipov et. al [10] propose the first electronic dictionary of Uzbek words' endings invariants for morphological segmentation pre-processing useful for neural machine translation.

The article [11] presents the algorithm of cosine similarity of Uzbek texts, based on TF-IDF to determine similarity. Another work on similarity in Uzbek, but this time on semantic similarity of words, a decent amount of work went on the creation and evaluation of a semantic evaluation dataset that possesses both similarity and relatedness scores **¡Error! No se encuentra el origen de la referencia.**.

## 3. Methodology

The scientific novelty of the methodology used in this work can be shown as follows:
- The creation of probability distributions law based on TF-IDF scores of unique words;
- Thorough investigation of numerical characteristics of the probability of unique words;
- Better evaluation of the stop words detection method's accuracy;

Summarising the automatic detection of the position of stop words in given Uzbek texts.

In our previous work[8], we proposed the usage of TF-IDF [15] to automatically extract stop words from a corpus of documents. The stop words are discovered based on the Term Frequency Inverse Document Frequency – TF-IDF. The number of times a word occurs in a text is defined by Term Frequency -- TF. Inverse Document Frequency -- IDF is defined as the number of texts (documents) being viewed and the presence of a given word in chosen texts (documents). TF-IDF is one of the popular methods of knowledge discovery.

Madatov et. al [8] propose the usage of TF-IDF [15] to automatically extract stop words from a corpus of documents. The stop words are discovered based on the frequency of the word and the frequency of the inverse document Term Frequency – Inverse Document Frequency – TF-IDF. The number of times a word occurs in a text is defined by Term Frequency -- TF. Inverse Document Frequency -- IDF is defined as the number of texts (documents) being viewed and the presence of a given word in chosen texts (documents). TF-IDF is one of the popular methods of knowledge discovery.

## 3.1. Probability distribution

In order to determine the position of the stop words throughout the school corpus, we investigate the probability distribution law of TF-IDF scores of stop words.

**Word weight and its probability.** Select a word $a_i; i \in [1..47165]$ from the set of unique words extracted from a corpus. For future references these two assumptions are valid: a word represents a unique word from a corpus and a corpus represents the "School corpus" presented in our previous work [4]. For every $a_i$ calculate average TF-IDF($a_i$), called the weight of $a_i$ and denoted as $w_i$. It is known that $w_i$ is not the probability of the word $a_i$.

The probability $p_i$ of *the word* $a_i$ can be calculated using the following formula: $p_i = w_i / \sum w_i$. We match $p_i$ for each $a_i$ word. Now $\sum p_i = 1$.

**The probability density function.** Suppose unique words are distributed independently in the total corpus. In that case, word $a_i$ can be applied multiple times. In order to escape repeating the word $a_i$ We consider only the first appearance of this word. For each word $a_i$ observe $i$ as a random variable. As the probability density function of the unique words, we get the following function:
$$f(i) = p_i$$
$f(i)$ can be considered as the probability density function of word $a_i$.

In the Cartesian coordinate plane, observe $i$ on the OX axis and observe $p_i$ along the OY axis. Figure 1 presents the described observations extracted from the "School corpus". We need it to observe the position of stop words along with the corpus.

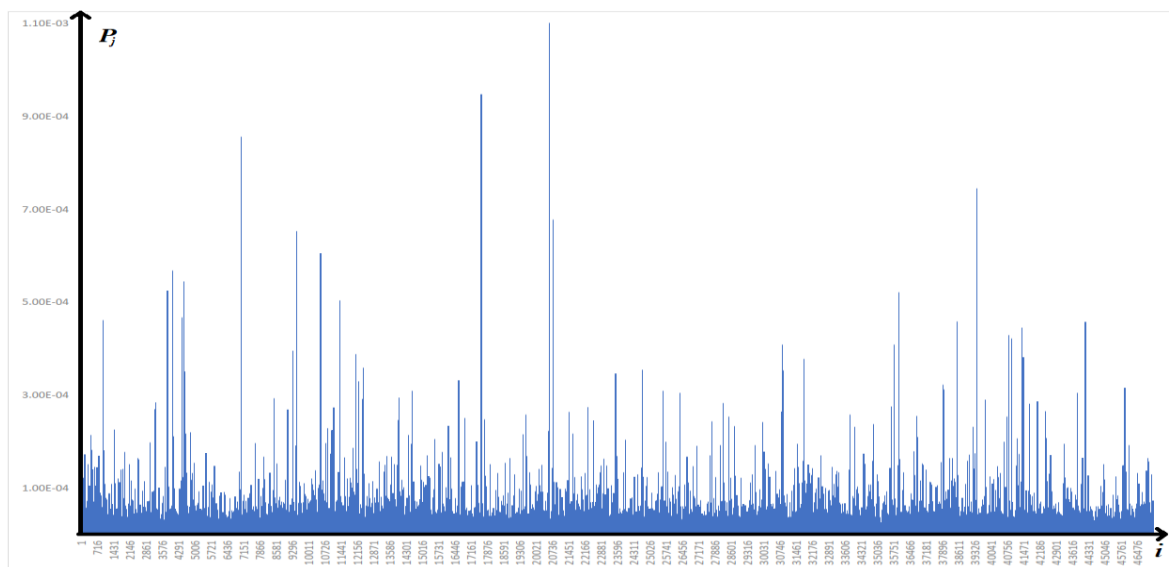

*Figure 1. The probability density function of unique words. The X-axis represents the index number of words, while the Y-axis shows the probability score.*

## 3.2. Numerical characteristics of the probability

This section presents numerical characteristics of the probability of unique words. They are calculated by the following formulas:

$E = \sum i \cdot p_i$ – the mathematical expectation of the unique words

$D = \sum (i - E)^2 \cdot p_i$ – dispersion of the unique words

$\sigma = \sqrt{D}$ – standard deviation of the unique words

$E_k = \sum p_i \cdot i^k$ – $k-th\ raw\ moments$ of the unique words

$\mu_3 = E_3 - 3 \cdot E_1 \cdot E_2 + 2 \cdot E_1^3$ — third central moment of the unique words

$A_s = \mu_3/\sigma^3$ — The asymmetry of the theoretical distribution

The described values extracted from the corpus are presented in Table 1.

*Table 1: Basic statistical properties extracted from the corpus.*

| $E$ | $D$ | $\sigma$ | $\sigma^3$ | $E_1$ | $E_2$ | $E_3$ | $\mu_3$ | $A_s$ |
|---|---|---|---|---|---|---|---|---|
| 23310,74 | 23310,74 | 13623,72 | 2,52864E+12 | 23310,74 | 728996416,52 | 25687931167881,50 | 41266663785,91 | 0,163 |

The variety of words increases gradually with grades in the school literature. It means that the probability density function of unique words is not symmetrical. One may predict it without a mathematical way. However, mathematically, the data in Table 1, especially, $A_s > o$, confirms that the probability density function is asymmetric.

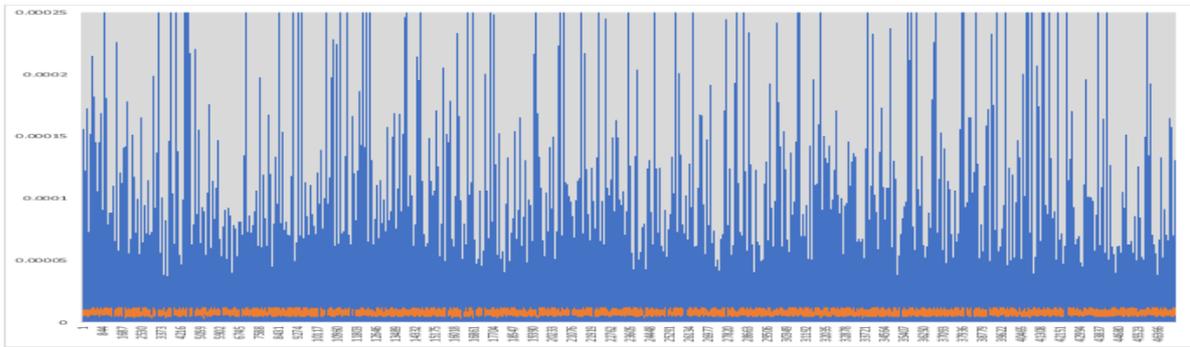

*Figure 2. The probability density function of unique words with stop words. The orange dots indicate the positions of stop words along with the corpus.*

The stop words are distributed along the axis (not grouped at one part of the axis); represented by orange dots in Figure 2.

### 3.3. Evaluation using a sub-corpus

This section presents the probability density function of unique words of selected work from the corpus. Each book from the corpus is devoted to one topic.

The prediction: Every book consists of the culmination part of the topic, the rest can be stop words. That is why we investigated just one book.

A random book was selected from the range of 25 books (in the corpus): 11[th] class literature. The book consists of 12837 unique words. The same process that was presented in Section 3.2 was applied to just the selected part of the corpus in order to create the probability density function of unique words. Figure 3 shows the probability density function of 11[th] class literature unique words.

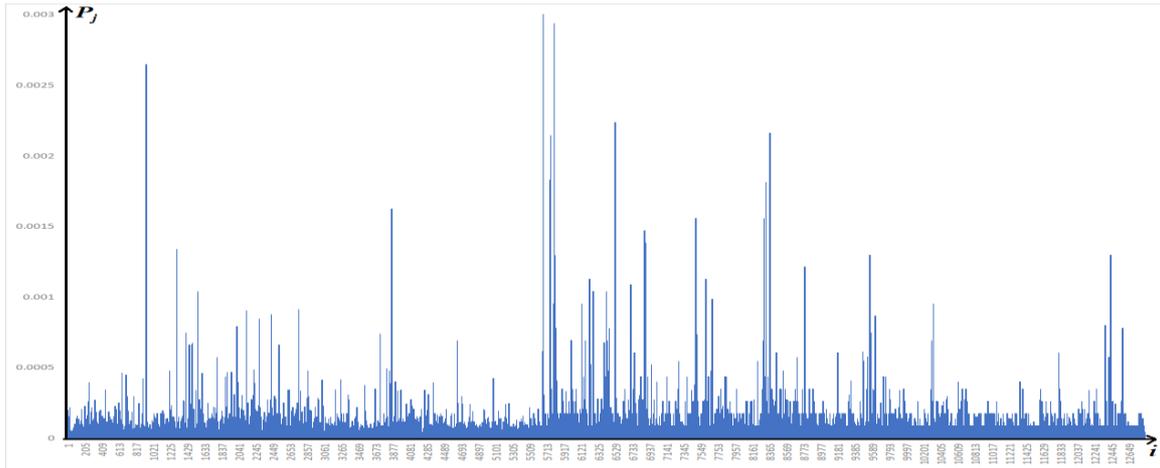

*Figure 3: probability density function of 11th class literature unique words*

Mathematical analysis of the distribution is presented in Table 2.
*Table 2: Distribution analysis of the selected single book*

| $E$ | $D$ | $\sigma$ | $\sigma^3$ | $E_1$ | $E_2$ | $E_3$ | $\mu_3$ | $A_s$ |
|---|---|---|---|---|---|---|---|---|
| 7076,623 | 11981425 | 3461,41 | 414472396507 | 7076,623 | 602060020 | 598084106956 | -10667328016 | -0,251 |

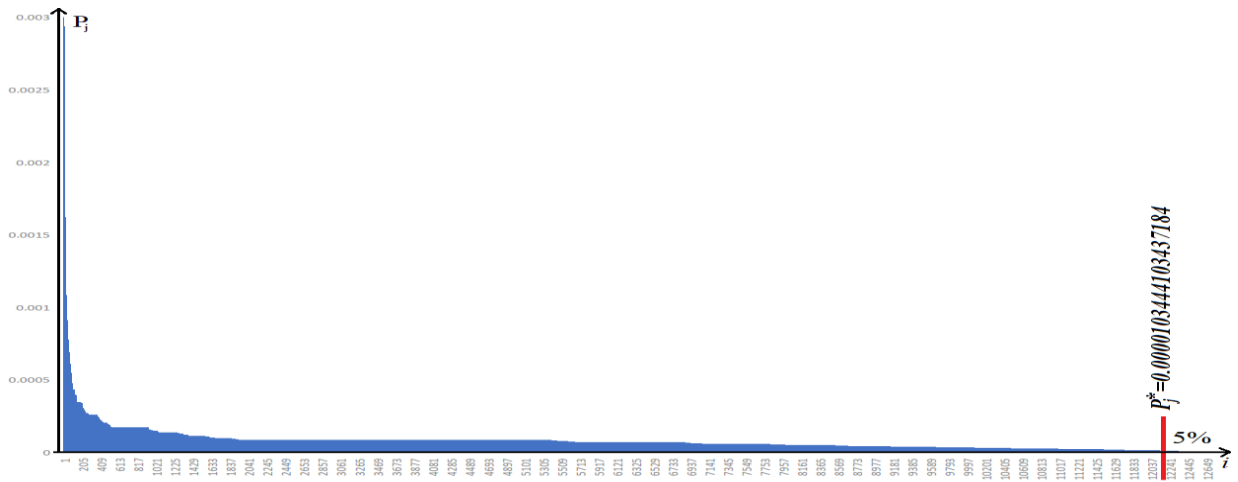

*Figure 4: Unique words from part of the corpus sorted by probability, lowest 5% are candidates for stop words*

We obtain Figure 4 by the rule of stop words detection method, as mentioned in [4].

$A_s < 0$ means that the probability density function is asymmetric.

The values were sorted in descending order and the lowest 5 percent of them are candidates for stop words. Figure 4 graphically represents the process, words with probability less than $p_{i^*}$ are candidates to be a stop word ($p_{i^*} = 0,00001034371184$).

The number of these candidates is 642. 85,8% of these words is located outside of the interval ($E - \sigma, E + \sigma$). On the left side of the interval there are 545 stop words and on the right side are 6 stop words. The same facts can be observed graphically on Figure 5 (Taking into the account the numerical characteristics of 5% words of selected work and comparing Figure 3 and figure 4 we detected their position along with the text).

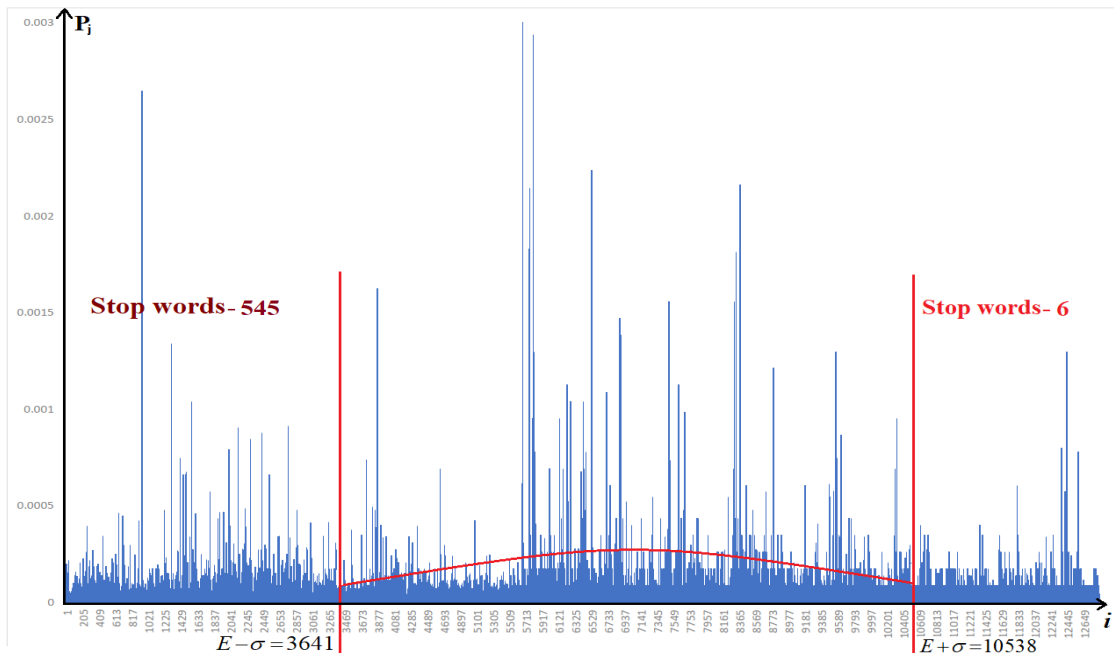
Figure 5: 85,8% of the stop word candidates are indeed located outside of the(E-σ,E+σ) interval

## 4. Evaluation results

The accuracy of the presented method if confirmed using the following reasoning:
Let suppose hypothesis
H0: Stop words of the selected document (11th class literature) are located outside of the interval (E-σ,E+σ);
and alternative hypothesis
H1: Stop words of the selected document (11th class literature) are located inside of the interval (E-σ,E+σ).

The critical value – Z (Z-score or Standard score) is obtained using this Equation:
$Z = \frac{\underline{X} - E}{\sigma/\sqrt{N}}$.; where N=12837, $\underline{X}$=6419, E=7076.62, $\sigma = 3461.419$.
In the presented task |Z|≈21,526. Z is located on the left side of E-σ, meaning there is no reason to reject the null hypothesis.
   This is the basis for rejecting the H1 hypothesis.

## 5. Conclusions and further work

Throughout the work performed in this paper, we presented a natural extension of the already presented previous research of the automatic detection of stop words in the Uzbek language [4] and the main focus of the analysis was twofold: a) a probability distributions model of the observed text and b) the accuracy of the detection method for stop words.
From all theoretical investigations from previous sections, it can be concluded that, for a single genre, the majority of stopwords have the following nature:
if $A_s < 0$, are located at the beginning parts of the text;
   if $A_s > 0$, are located at the ending of the text;
   if $A_s = 0$, are located at the beginning at the ending part of the text.

In future works, we would like to use the results of this article as the basis for automatically extracting keywords and automatically extracting the abstract of a given text.

## 6. Acknowledgements

The authors gratefully acknowledge the European Commission for funding the InnoRenew CoE project (Grant Agreement $\#$739574) under the Horizon2020 Widespread-Teaming program and the Republic of Slovenia (Investment funding of the Republic of Slovenia and the European Union of the European Regional Development Fund).

## 7. Conclusion

The paper presents a natural extension of the already presented research of automatic detection of stop words in Uzbek language[8] and presents two goals: a) a probability distributions model of the observed text and b) the accuracy of the detection method for stop words.
   a) The probability density is defined and later used to observe the accuracy of the automatic method for extraction of stop words of Uzbek language.
   b) The accuracy of the method that is presented in Section **¡Error! No se encuentra el origen de la referencia.**.

From this fact it can be concluded that, for a single genre, more of the stop words for texts:
if $A_s < 0$, are located at the beginning parts of the text;
if $A_s > 0$, are located at the ending of the text;
if $A_s = 0$, are located at the beginning at the ending part of the text.
Further we use this result in the process of automatically extracting keywords from the given text and automatically extracting the annotation of the given text.